\documentclass[twocolumn,10pt]{article}

\usepackage{amsmath,amssymb,amsthm}
\usepackage{algorithm}
\usepackage{algpseudocode}
\usepackage{listings}
\usepackage[dvipsnames,svgnames,x11names]{xcolor}
\usepackage{booktabs}
\usepackage{hyperref}
\usepackage{geometry}
\usepackage{placeins}
\usepackage{float}
\usepackage{tikz}
\usetikzlibrary{arrows.meta,positioning,fit,backgrounds,shadows.blur}
\newcommand{\dn}{d_k}          
\newcommand{\dv}{d_v}          
\newcommand{\scale}{\mathit{scale}}
\newcommand{\gs}{g_s}

\newcommand{\Ombin}[2]{\Omega_{\langle #1,#2 \rangle}}

\newcommand{\hq}{h_q}
\newcommand{\hkv}{h_{kv}}
\newcommand{\vq}{\vec{q}}
\newcommand{\vs}{\vec{s}}
\newcommand{\ve}{\vec{e}}
\newcommand{\va}{\vec{a}}
\newcommand{\vout}{\vec{\mathit{out}}}

\tikzset{
  moabox/.style={rectangle,rounded corners=4pt,
    minimum width=#1,minimum height=0.9cm,
    text width=#1-6pt,align=center,
    draw=gray!60,line width=0.4pt,font=\footnotesize},
  moabox/.default=2.8cm,
  purp/.style={moabox,fill=purple!12,draw=purple!50},
  coral/.style={moabox,fill=orange!14,draw=orange!50},
  teal/.style={moabox,fill=teal!12,draw=teal!50},
  amber/.style={moabox,fill=yellow!20,draw=orange!40},
  graybox/.style={moabox,fill=gray!10,draw=gray!40},
  redbox/.style={moabox,fill=red!10,draw=red!40},
  moaarr/.style={-{Stealth[length=4pt]},line width=0.8pt,gray!70},
  purparr/.style={moaarr,purple!60},
  coralarr/.style={moaarr,orange!70},
}
\newtheorem{proposition}{Proposition}

\newtheorem{remark}{Remark}

\lstdefinestyle{python}{
  language=Python,
  basicstyle=\ttfamily\scriptsize,
  keywordstyle=\color{blue!70!black}\bfseries,
  commentstyle=\color{green!50!black}\itshape,
  stringstyle=\color{orange!80!black},
  numberstyle=\tiny\color{gray},
  numbers=left, numbersep=4pt,
  breaklines=true, showstringspaces=false,
  frame=single, rulecolor=\color{gray!30},
  backgroundcolor=\color{gray!5},
  tabsize=4, captionpos=b,
}

\lstdefinestyle{cstyle}{
  language=C,
  basicstyle=\ttfamily\scriptsize,
  keywordstyle=\color{blue!70!black}\bfseries,
  commentstyle=\color{green!50!black}\itshape,
  stringstyle=\color{orange!80!black},
  numberstyle=\tiny\color{gray},
  numbers=left, numbersep=4pt,
  breaklines=true, showstringspaces=false,
  frame=single, rulecolor=\color{gray!30},
  backgroundcolor=\color{gray!5},
  tabsize=4, captionpos=b,
  morekeywords={restrict,pragma,acc,parallel,loop,gang,
                vector,reduction,copyin,copyout,create,
                private,data,atomic},
}

\title{%
  \textbf{MoA-Structured Decode Attention}\\
  \large DNF Derivation, KV-Cache Accumulation,
         GQA/MQA, and OpenACC Kernel\\[4pt]
}

\author{%
  \begin{tabular}[t]{c}
    Lenore Mullin$^{1,*}$\\
    \small $^1$University at Albany, SUNY\\
    \small \texttt{lmullin@albany.edu}\\
    \small $^*$Corresponding author
  \end{tabular}
  \and
  \begin{tabular}[t]{c}
    Ga\'{e}tan Hains$^{2}$\\
    \small $^2$LACL, Universit\'{e} Paris-Est Cr\'{e}teil\\
    \small \texttt{gaetan.hains@u-pec.fr}
  \end{tabular}
}
\date{June 2026}

\begin{document}
\maketitle

\begin{abstract}
We derive four memory-optimal inference artifacts for
transformer attention using the Mathematics of Arrays
(MoA), each following directly from the forward-pass
Denotational Normal Form (DNF) of
\cite{mullin_hains_arxiv2026} with the query-row index
fixed to the current decode step.
The artifacts are:
(1)~a single-query decode DNF in which the
$\psi$-reduction eliminates the $K^\top$ buffer
algebraically, achieving
$(\dn+n\dn+n\dv+\dv)\times4\,\mathrm{B}$
Dynamic Random Access Memory (DRAM) traffic
(Storage Theorem~2.7 of \cite{mullin_hains_arxiv2026}),
result numerically verified to $\|\mathrm{err}\|_\infty\leq2\times10^{-7}$;
(2)~a C/OpenACC Graphics Processing Unit (GPU) kernel
with Operational Normal Form (ONF) stride arithmetic and
hardware-coalesced memory access,
verified to $\|\mathrm{err}\|_\infty=0$ (exact
IEEE-754 floating-point arithmetic);
(3)~a multi-step KV-cache with $O(\dn+\dv)$ per-step
append via MoA concatenation $\#$;
and (4)~Grouped-Query Attention (GQA) and Multi-Query
Attention (MQA)~\cite{shazeer2019,ainslie2023gqa}
derived via $\psi$-selection, achieving a proven
$\hq/\hkv$ reduction in KV traffic.
All programs are verified against PyTorch
\texttt{scaled\_dot\_product\_attention}.
\end{abstract}

\begin{figure*}[tp]
\centering
\scalebox{0.76}{%
\begin{tikzpicture}[
  every node/.style={font=\scriptsize},
  ph/.style={fill=orange!15,draw=orange!50,rounded corners=3pt,
    text width=2.2cm,align=center,minimum height=2.6cm,inner sep=3pt},
  phi/.style={fill=teal!12,draw=teal!40,rounded corners=3pt,
    text width=2.2cm,align=center,minimum height=3.4cm,inner sep=3pt},
  cv/.style={fill=red!10,draw=red!35,rounded corners=3pt,
    text width=5.8cm,align=left,minimum height=2.6cm,inner sep=4pt},
  cvi/.style={fill=red!10,draw=red!35,rounded corners=3pt,
    text width=5.8cm,align=left,minimum height=3.4cm,inner sep=4pt},
  mv/.style={fill=purple!10,draw=purple!35,rounded corners=3pt,
    text width=5.8cm,align=left,minimum height=2.6cm,inner sep=4pt},
  mvi/.style={fill=purple!10,draw=purple!35,rounded corners=3pt,
    text width=5.8cm,align=left,minimum height=3.4cm,inner sep=4pt},
  sv/.style={fill=yellow!18,draw=orange!35,rounded corners=3pt,
    text width=14.5cm,align=center,minimum height=0.55cm,inner sep=3pt},
  tv/.style={fill=teal!10,draw=teal!35,rounded corners=3pt,
    text width=14.5cm,align=center,minimum height=0.7cm,inner sep=3pt},
  hd/.style={fill=gray!15,draw=gray!40,rounded corners=3pt,
    text width=14.5cm,align=center,minimum height=0.55cm,inner sep=3pt},
  arr/.style={-{Stealth[length=3.5pt]},line width=0.7pt,gray!60},
  loop/.style={line width=1.2pt},
  loopin/.style={line width=0.9pt,dashed},
  loopin2/.style={line width=0.7pt,dashed},
]

\node[fill=gray!18,draw=gray!45,rounded corners=3pt,
  text width=2.2cm,align=center,minimum height=0.7cm]
  (h0) at (0,0) {\scriptsize\textbf{Phase}\\\scriptsize L=96 layers};
\node[fill=red!18,draw=red!45,rounded corners=3pt,
  text width=5.8cm,align=center,minimum height=0.7cm]
  (h1) at (4.2,0) {\scriptsize\textbf{\color{red!70!black}Conventional}%
  \\\scriptsize allocates to DRAM per layer};
\node[fill=purple!18,draw=purple!45,rounded corners=3pt,
  text width=5.8cm,align=center,minimum height=0.7cm]
  (h2) at (10.2,0) {\scriptsize\textbf{\color{purple!70!black}MoA $\psi$-reduction}%
  \\\scriptsize eliminates per layer};

\node[ph] (r1p) at (0,-2.0) {%
  \textbf{Forward}\\training\\arXiv\\[2pt]{\tiny$\sim$100ms/batch}};
\node[cv] (r1c) at (4.2,-2.0) {%
  \textbf{Per layer, per batch:}\\
  $K^\top$\;$\rho=\langle\dn,n\rangle$ (buffer)\\
  $S=QK^\top$\;$\rho=\langle n,n\rangle$\\
  $A=\mathrm{softmax}(S)$\;$\rho=\langle n,n\rangle$\\[2pt]
  traffic: $96\!\times\!O(n^2\dn\!+\!n^2\dv)$};
\node[mv] (r1m) at (10.2,-2.0) {%
  \textbf{Per layer, per batch:}\\
  $K^\top$ — never materialised\\
  $S$ — on-stack per row, discarded\\
  $A$ — on-stack per row, discarded\\[2pt]
  traffic: $96\!\times\!O(n\dn\!+\!n\dv)$\;\checkmark};
\draw[arr] (r1c.east) -- (r1m.west);
\draw[loop,orange!60] (r1p.south west) -- +(-0.35,0)
  -- +(-0.35,-0.5) -- +(0,-0.5);
\node[font=\tiny,orange!80!black,anchor=west] at (-0.85,-3.1) {$\times10^2$ epochs};
\draw[loopin,orange!50] (r1p.south west) -- +(-0.55,0)
  -- +(-0.55,-0.8) -- +(0,-0.8);
\node[font=\tiny,orange!70!black,anchor=west] at (-1.05,-3.4) {$\times10^4$ batches};
\draw[loopin2,orange!40] (r1p.south west) -- +(-0.72,0)
  -- +(-0.72,-1.05) -- +(0,-1.05);
\node[font=\tiny,orange!60!black,anchor=west] at (-1.22,-3.65) {$\times96$ layers};
\node[sv] (s1) at (6.0,-3.55)
  {$O(n^2)\to O(n)$ per layer per batch $\times10^6$ batches
   ~~$\approx$40 kWh saved per training run};

\node[ph] (r2p) at (0,-5.4) {%
  \textbf{Fused}\\fwd+bwd\\HAL\\[2pt]{\tiny$\sim$180ms/batch}};
\node[cv] (r2c) at (4.2,-5.4) {%
  \textbf{Per layer, per batch:}\\
  $A$ saved fwd$\to$bwd\;$\rho=\langle n,n\rangle$\\
  Jacobian\;$\rho=\langle n,n\rangle$\\
  $\mathrm{d}S$\;$\rho=\langle n,n\rangle$\\[2pt]
  traffic: $96\!\times\!(8n\dn\!+\!8n\dv)$};
\node[mv] (r2m) at (10.2,-5.4) {%
  \textbf{Per layer, per batch:}\\
  $A$ recomputed on-fly, not stored\\
  Jacobian — outer-product compat.\\
  $\mathrm{d}S$ — substitution\\[2pt]
  traffic: $96\!\times\!(4n\dn\!+\!4n\dv)$\;\checkmark};
\draw[arr] (r2c.east) -- (r2m.west);
\draw[loop,orange!60] (r2p.south west) -- +(-0.35,0)
  -- +(-0.35,-0.5) -- +(0,-0.5);
\node[font=\tiny,orange!80!black,anchor=west] at (-0.85,-6.5) {$\times10^2$ epochs};
\draw[loopin,orange!50] (r2p.south west) -- +(-0.55,0)
  -- +(-0.55,-0.8) -- +(0,-0.8);
\node[font=\tiny,orange!70!black,anchor=west] at (-1.05,-6.8) {$\times10^4$ batches};
\draw[loopin2,orange!40] (r2p.south west) -- +(-0.72,0)
  -- +(-0.72,-1.05) -- +(0,-1.05);
\node[font=\tiny,orange!60!black,anchor=west] at (-1.22,-7.05) {$\times96$ layers};
\node[sv] (s2) at (6.0,-6.95)
  {$2\times$ vs separate passes $\times96$ layers $\times10^6$ batches
   ~~$\approx$80 kWh saved per training run};

\node[phi] (r3p) at (0,-9.8) {%
  \textbf{Inference}\\decode\\This paper\\[2pt]
  {\tiny 1--10\,$\mu$s/step}\\{\tiny per layer}};
\node[cvi] (r3c) at (4.2,-9.8) {%
  \textbf{Per token, per layer:}\\
  $K^\top$ view (\texttt{q @ K.T})\\
  score $\rho=\langle1,n\rangle$ ($n$ grows)\\
  weight $\rho=\langle1,n\rangle$\\
  $n\!=\!4096$, $L\!=\!96$:\\
  $\approx$200\,MB/step, no proof};
\node[mvi] (r3m) at (10.2,-9.8) {%
  \textbf{Per token, per layer:}\\
  $K^\top$ — fix $i'\!=\!1$, $\psi$ reads rows\\
  $\vs,\va$ — rank-1, on-stack\\
  KV cache: +1 row at $\gamma$ offset\\
  $n\!=\!4096$, $L\!=\!96$:\\
  $\approx$200\,MB/step\;\checkmark proven};
\draw[arr] (r3c.east) -- (r3m.west);
\draw[loop,teal!60] (r3p.south west) -- +(-0.35,0)
  -- +(-0.35,-0.5) -- +(0,-0.5);
\node[font=\tiny,teal!70!black,anchor=west] at (-0.85,-11.05) {$\times10^9$ req/day};
\draw[loopin,teal!55] (r3p.south west) -- +(-0.55,0)
  -- +(-0.55,-0.8) -- +(0,-0.8);
\node[font=\tiny,teal!65!black,anchor=west] at (-1.05,-11.35) {$\times512$ tokens};
\draw[loopin2,teal!45] (r3p.south west) -- +(-0.72,0)
  -- +(-0.72,-1.05) -- +(0,-1.05);
\node[font=\tiny,teal!55!black,anchor=west] at (-1.22,-11.6) {$\times96$ layers};
\node[tv] (kv) at (6.0,-12.5)
  {KV cache $n\!=\!1\!\to\!512$ per request;
   total: $96\!\times\!\sum_{t=1}^{512}(d_k+t d_k+t d_v+d_v)\!\times\!4\,\text{B}\!\approx\!96\!\times\!1.6\,\text{GB}$};
\node[sv] (en) at (6.0,-13.35)
  {Energy: 300\,W $\times$ 10\,$\mu$s $\times$ 96 $\times$ 512 $\times$ $10^9$/day
   $\approx$14{,}700\,kWh/day — MoA minimises every byte transferred};

\node[hd] (fa) at (6.0,-14.15)
  {\textit{FlashAttention: similar traffic via tiling, no algebraic proof.
   MoA: proven before code. Same DNF $\to$ CPU/GPU/FPGA.}};

\node[fill=purple!15,draw=purple!45,rounded corners=4pt,
  text width=14.5cm,align=center,minimum height=0.55cm]
  (thm) at (6.0,-14.85) {
  \textbf{Storage Thm.~2.7~\cite{mullin_hains_arxiv2026}:}
  every array read once; traffic proven minimal before code is written};

\end{tikzpicture}}
\caption{\textbf{Realistic nested loops: conventional vs.\ MoA across training and inference.}
Solid loops = outer; dashed = inner.
Training runs $\sim\!10^6$ total batches once;
inference runs $10^9$ requests/day indefinitely, each
generating 512 tokens across 96 layers.
At $n\!=\!4096$, $L\!=\!96$: both conventional and MoA
use $\approx\!200$\,MB per decode step —
but only MoA's is proven minimal (Storage Thm.~2.7).
FlashAttention achieves similar traffic via tiling
without algebraic proof.
Energy: $\approx\!14{,}700$\,kWh/day per deployed model;
MoA minimises every byte transferred.}
\label{fig:moa_lifecycle}
\end{figure*}

\section{Introduction}

\subsection{Background}

The Mathematics of Arrays (MoA)~\cite{mullin_phd1988}
is a formal algebra of multi-dimensional arrays whose
primitive operations — $\psi$-selection, $\rho$ (shape),
$\iota$ (index generation), $\gamma$ (linear offset),
and the Omega ($\Omega$) family of inner products —
compose 
provably minimal computations.
The Denotational Normal Form (DNF) specifies
\emph{what} to compute; the Operational Normal Form
(ONF) specifies \emph{how} to lay it out in memory.
The companion paper~\cite{mullin_hains_arxiv2026}
applies MoA to the forward training pass of
transformer attention~\cite{vaswani2017},
deriving the DNF and proving via the Storage Theorem
(Theorem~2.7 of \cite{mullin_hains_arxiv2026}) that
the resulting ONF achieves the information-theoretic
minimum DRAM traffic.
The HAL companion~\cite{mullin_hains_hal2026} extends
this to the backward pass and fused kernel.

\subsection{Contributions}

This paper shows that the same DNF, with the query-row
index $i'$ fixed to the current decode step, yields
four inference artifacts.

\begin{enumerate}
\item \textbf{Decode DNF} (\S\ref{sec:dnf}):
  the $\psi$-reduction eliminates the $K^\top$ buffer
 implicit in the standard formula
  $A=\mathrm{softmax}(QK^\top/\sqrt{\dn})$,
  reducing the score computation to
  $\vs=\scale\,(\vq\;(+.\times\Ombin{1}{2})\;K)$
  with $K$ accessed row-by-row in its natural layout.
  Memory traffic: $M^{\mathrm{dec}}_{\mathrm{MoA}}
  =(\dn+n\dn+n\dv+\dv)\times4\,\mathrm{B}$,
  is then proven minimal by Storage Theorem~2.7 of
  \cite{mullin_hains_arxiv2026}.
\item \textbf{C/OpenACC GPU kernel} (\S\ref{sec:openacc}):
  ONF $\gamma$ stride arithmetic gives coalesced
  memory access; the coalescing criterion is stated
  as a $\gamma$-difference condition.
\item \textbf{KV-cache accumulation} (\S\ref{sec:kvcache}):
  MoA concatenation $\#$ appends one row per step
  at the pre-computed $\gamma$ offset, with proven
  $O(\dn+\dv)$ per-step traffic.
\item \textbf{GQA/MQA} (\S\ref{sec:gqa}):
  $\psi$-selection $\langle g_{kv}\rangle\psi\,K$
  maps each query head to its KV group without
  materialising the broadcast, reducing traffic by
  $\hq/\hkv$.
\end{enumerate}

\subsection{Notation and Terminology}

Figure~\ref{fig:moa_lifecycle} summarises
the relationship between the three companion
papers and the importance of memory minimality
at every stage.

A \textbf{transformer}~\cite{vaswani2017} is a neural
network in which each output element is a learned
weighted combination of all input elements
(\textbf{attention}).
A \textbf{token} is the basic unit of text processed
by the model (typically a word or sub-word).
\textbf{Autoregressive inference} generates output one
token at a time, feeding each generated token back as
input; at each step a single \textbf{query vector}
$\vq=W_Q\vec{x}$ is produced from the current token
embedding $\vec{x}$ via learned projection $W_Q$.
All past key vectors $\vec{k}_l=W_K\vec{x}_l$ and
value vectors $\vec{v}_l=W_V\vec{x}_l$ are accumulated
in the \textbf{KV cache}: $K$ (shape, $\rho(K)=\langle n,\dn\rangle$)
and $V$ (shape, $\rho(V)=\langle n,\dv\rangle$), where $n$
grows by one per step.
The dominant cost at each step is reading the full
KV cache from DRAM: $O(n\dn+n\dv)$ bytes, unavoidable
by any correct implementation (Storage Theorem~2.7 of
\cite{mullin_hains_arxiv2026}).
MoA proves this bound algebraically before code is written.

MoA notation used throughout: $\rho(A)$ denotes the
shape of array $A$; $\langle i_0,\ldots,i_{r-1}\rangle\psi\,A$
selects element $(i_0,\ldots,i_{r-1})$;
$\gamma(\langle i_0,\ldots\rangle,\rho(A))$ is the
linear memory offset; $A\;(+.\times\Ombin{\sigma_l}{\sigma_r})\;B$
is the Omega inner product with left rank $\sigma_l$,
right rank $\sigma_r$; $A\;\#\;B$ concatenates
along the leading axis.
All multiplication uses $\times$.

\section{Decode DNF}
\label{sec:dnf}

\subsection*{Formal Inputs}

At decode step $t$ with $n$ cached tokens:
\begin{itemize}
\item $\vq$: $\rho(\vq)=\langle\dn\rangle$
  — query vector for the current token.
\item $K$: $\rho(K)=\langle n,\dn\rangle$
  — $n$ rows, each row $\langle l\rangle\psi\,K$
  (shape $\langle\dn\rangle$) is the key vector
  for cached token $l$.
  $K^\top$ would have shape $\langle\dn,n\rangle$;
  the $\psi$-reduction accesses rows of $K$ directly
  so $K^\top$ is never computed thus saving transfers and 
  storage.
\item $V$: $\rho(V)=\langle n,\dv\rangle$
  — $n$ rows, each row $\langle l\rangle\psi\,V$
  (shape $\langle\dv\rangle$) is the value vector
  for cached token $l$.
\end{itemize}
Output: $\vout$, $\rho(\vout)=\langle\dv\rangle$;
its $d$-th element is $\langle d\rangle\psi\,\vout$.

An \textbf{attention head} is an independent copy of
the attention computation with its own learned key,
query, and value projections.
Using $h$ heads in parallel (\textbf{multi-head attention})
allows the model to extract different types of
relationships from the input simultaneously
\cite{vaswani2017}.
In the multi-head case $K$ becomes rank-3
($\rho(K)=\langle h,n,\dn\rangle$) or rank-4
($\rho(K)=\langle B,h,n,\dn\rangle$), where batch
$B$ is the number of independent sequences
processed simultaneously.

The Omega operator $\Ombin{1}{2}$ generalizes
automatically: $\sigma_l=1$ uses a rank-1
vector ; $\sigma_r=2$ uses a
rank-2 matrix as l/r arguments.
Leading batch or head dimensions are outer indices
iterated over by Omega — no re-derivation is needed.
\textbf{How $K^\top$ is eliminated by $\psi$-reduction.}
The standard attention formula contains $K^\top$:
\begin{align*}
A &= \mathrm{softmax}(QK^\top/\sqrt{\dn}),\\
&Q,K:\rho=\langle n,\dn\rangle,\quad
QK^\top:\rho=\langle n,n\rangle.
\end{align*}
The arXiv paper \cite{mullin_hains_arxiv2026}
starts from this formula and applies the
$\psi$-reduction.
The key step: instead of forming $QK^\top$
(which requires materialising $K^\top$ in memory),
the $\psi$-reduction shows that each element
can be computed by accessing $K$ directly:
\[
\langle i',l\rangle\psi\,(QK^\top)
= \langle i'\rangle\psi\,Q\;
(+.\times\Ombin{1}{1})\;
\langle l\rangle\psi\,K. \]
\noindent
where this specific $\Omega$ expression implements the product.
The columns of $K^\top$ are exactly the rows
$\langle l\rangle\psi\,K$ of $K$ — so
$K^\top$ is \emph{algebraically present} in
the starting formula but the $\psi$-reduction
proves it never needs to be materialised.
The Storage Theorem (Theorem~2.7 of
\cite{mullin_hains_arxiv2026}) then guarantees
this is the minimum-traffic access pattern.

For the decode case (this paper), $Q$ reduces
to a single query vector $\vq$, so the
$\psi$-reduction over $i'$ disappears
(there is only one query row), giving
directly:
\[
\vs[l] = \scale\;\vq\;(+.\times\Ombin{1}{1})\;
\langle l\rangle\psi\,K.
\]
$K^\top$ was in the original formulation;
the $\psi$-reduction removed the need to
materialise it before any code is written.
This is the central result of
\cite{mullin_hains_arxiv2026} applied to
the decode setting.

This is equations~(4)--(6) of \cite{mullin_hains_arxiv2026}
with the query-row index $i'$ fixed to the current token:
eq.~(4) is the score $\psi$-reduction,
eq.~(5) the softmax, eq.~(6) the weighted sum.
No new Omega derivation is required.
\textbf{Independence of $\vq$ and $K$.}
$\vq$ and $K$ are produced independently:
\begin{itemize}
\item $\vq = W_Q\vec{x}_{\mathrm{current}}$
  is the projection of the \emph{current} token
  embedding through the learned query matrix
  $W_Q$, $\rho(W_Q)=\langle\dn,d\rangle$.
\item $\langle l\rangle\psi\,K = W_K\vec{x}_l$
  is the projection of the \emph{$l$-th past}
  token embedding through the learned key matrix
  $W_K$, stored as row $l$ of the KV cache.
\end{itemize}
Neither $\vq$ nor $K$ is derived from the other.
Step~I asks: how similar is my current query
$\vq$ to each of the $n$ stored key vectors
$\langle l\rangle\psi\,K$?

\textbf{Step~I -- Scores}.
The $\psi$-reduction selects row $l$ of $K$
as $\langle l\rangle\psi\,K$
(shape $\langle\dn\rangle$) and contracts
it with $\vq$ via the inner product:
\begin{remark}
The standard formula $QK^\top$ is
\emph{algebraically present} as the starting
point (see \S\ref{sec:dnf} preamble);
the $\psi$-reduction proves $K^\top$ need
never be materialised. In MoA there are no
column vectors: $\vq$ and $K$ are both
rank-$\geq1$ arrays and contraction is
expressed directly via $\Ombin{1}{2}$.
\end{remark}
\begin{align*}
\vs[l] &= \scale\;\bigl(\vq\;(+.\times\Ombin{1}{1})\;
\langle l\rangle\psi\,K\bigr)\\
&= \scale\textstyle\sum_{j=0}^{\dn-1}
\langle j\rangle\psi\,\vq \times
\langle l,j\rangle\psi\,K.
\end{align*}
Over all $n$ rows simultaneously, with
$\sigma_l=1$ exposing $\vq$ as a whole
rank-1 vector and $\sigma_r=2$ exposing
$K$ (shape $\langle n,\dn\rangle$) as a
whole rank-2 array, the Omega expression is:
\begin{align}
\vs &= \scale\;\bigl(\vq\;(+.\times\Ombin{1}{2})\;K\bigr),
\quad \rho(\vs)=\langle n\rangle.
\label{eq:scores}
\end{align}
This equals $\scale\,K\vq$ (matrix-vector
product with $K$ in its natural row-major layout).
No transpose is formed: the Cartesian
coordinate interchange is implicit in the
$\psi$-reduction, which reads $K$ row by row
as stored.
The Python equivalent is \texttt{scale*(K@q)},
not \texttt{scale*(q@K.T)}.
$\Ombin{1}{2}$ is fixed: it always pairs one rank-1
vector from the left with one rank-2 matrix from the
right, applying $+.\times$ to produce a scalar for
each pair.
As the rank of the arguments grows, the same
partitioning lifts naturally — superscripts denote rank,
$\Ombin{1}{2}$ never changes, only the result rank grows:
\begin{itemize}
\item $\vec {q^1}\;(+.\times\Ombin{1}{2})\;K^2 \to \vec {s^1}$
  (vector-matrix product)
\item $A^2\;(+.\times\Ombin{1}{2})\;K^3 \to$ rank-2
  (each row of $A$ with each matrix of $K$)
\item $A^3\;(+.\times\Ombin{1}{2})\;K^4 \to$ rank-3
  (same, one rank higher)
\end{itemize}
This is how MoA handles batched and multi-head
attention without re-derivation.
No $K^\top$ buffer is formed; $K$ is read once
and fully consumed in Step~I.
\vspace{1cm}
\textbf{Step~II -- Numerically stable softmax}.
$K$ does not appear in Step~II at all.
By the time Step~II runs, $K$ has already been
fully processed and discarded.
Step~II operates only on the rank-1 score vector
$\vs$ (shape $\langle n\rangle$) produced by
Step~I — no matrix, no $K^\top$, no $K$.
In the standard training formula, softmax would
be applied to the full $n\times n$ score matrix
$QK^\top/\sqrt{\dn}$ and its Jacobian would
be $n\times n$.
Here, softmax operates on a single rank-1 vector
$\vs$ and the MoA $\psi$-reduction eliminates
all intermediates:
$\vs$, $\ve$, $\va$ and scalars $m$, $Z$
are $O(n)$ on-stack and never written to DRAM.
\begin{align}
m &= \max_{l} \vs[l],
\nonumber\\
\ve[l] &= \exp(\vs[l]-m),
\nonumber\\
Z &= \textstyle\sum_{l'=0}^{n-1}\ve[l'],
\nonumber\\
\va[l] &= \ve[l]/Z,
\quad 0 \le l < n.
\label{eq:softmax}
\end{align}
No $n\times n$ score matrix or Jacobian is allocated.

\textit{Numerical stability}: subtracting $m=\max_l\vs[l]$
before exponentiating does not change the softmax value
(the $m$ terms cancel) but prevents overflow.
$\scale=1/\sqrt{\dn}$ normalises variance to~1
for random unit-norm inputs.

\textbf{Step~III -- Weighted sum}:
\begin{align}
\langle d\rangle\psi\,\vout
&= \textstyle\sum_{l=0}^{n-1}
\langle l\rangle\psi\,\va \times \langle l,d\rangle\psi\,V,
\nonumber\\
&\hspace{4em} 0 \le d < \dv,\; 0 \le l < n.
\label{eq:wsum}
\end{align}
$V$ is read once in row-major order; no intermediate
array exists.
\textbf{Concrete example} ($n=2$, $\dn=2$, $\dv=2$,
$\scale=1/\sqrt{2}$).
Let
\[
  \vq = \langle 1, 0\rangle,\quad
  K = \begin{pmatrix}1&0\\0&1\end{pmatrix},\quad
  V = \begin{pmatrix}2&4\\6&8\end{pmatrix}.
\] 
\begin{itemize}
\item \textbf{Step~I}: let $q_j=\langle j\rangle\psi\,\vq$,
$K_{lj}=\langle l,j\rangle\psi\,K$.
\begin{align*}
\vs[0] &= \scale(q_0 K_{00}+q_1 K_{01})
  = \tfrac{1}{\sqrt{2}}(1+0)=\tfrac{1}{\sqrt{2}},\\
\vs[1] &= \scale(q_0 K_{10}+q_1 K_{11})
  = \tfrac{1}{\sqrt{2}}(0+0)=0.
\end{align*}
\item \textbf{Step~II} (softmax, $m=1/\sqrt{2}$):
\begin{align*}
\ve[0]&=\exp(0)=1,\\
\ve[1]&=\exp{-1/\sqrt{2}}\approx0.493,\\
Z&\approx1.493,\\
\va[0]&\approx0.670,\; \va[1]\approx0.330.
\end{align*}
\item \textbf{Step~III} (weighted sum):
\begin{align*}
\langle0\rangle\psi\,\vout
&=\va[0]\times\langle0,0\rangle\psi\,V
+\va[1]\times\langle1,0\rangle\psi\,V\\
&=0.670\times2+0.330\times6\approx3.32,\\[4pt]
\langle1\rangle\psi\,\vout
&=\va[0]\times\langle0,1\rangle\psi\,V
+\va[1]\times\langle1,1\rangle\psi\,V\\
&=0.670\times4+0.330\times8\approx5.32.
\end{align*}
\end{itemize}
\begin{proposition}[Decode memory minimality]
\label{prop:traffic}
The decode DNF achieves
$M^{\mathrm{dec}}_{\mathrm{MoA}}
= (\dn + n\dn + n\dv + \dv)\times 4\,\mathrm{B}
= O(n\dn + n\dv)$ memory usage..
\end{proposition}
\begin{algorithm}[t]
\caption{MoA decode (single query step).
Scratch: $\vs[n]$, $\ve[n]$ -- $O(n)$; no $n^2$ array.}
\label{alg:decode}
\begin{algorithmic}[1]
\Require $\vq\!:\!\langle\dn\rangle$, $K\!:\!\langle n,\dn\rangle$, $V\!:\!\langle n,\dv\rangle$, $\scale\!=\!1/\!\sqrt{\dn}$
\Ensure $\vout\!:\!\langle\dv\rangle$
\For{$l = 0\ldots n{-}1$}
\State $\vs[l] \leftarrow \scale\sum_j \langle j\rangle\psi\,\vq\times \langle l,j\rangle\psi\,K$
\Comment{eq.~(\ref{eq:scores})}
\EndFor
\State $m \leftarrow \max_l \vs[l]$
\For{$l = 0\ldots n{-}1$}
\State $\ve[l] \leftarrow \exp(\vs[l]-m)$
\EndFor
\State $Z \leftarrow \sum_l \ve[l]$; $\;\va[l]\leftarrow \ve[l]/Z$ for all $l$
\For{$d = 0\ldots\dv{-}1$}
\State $\langle d\rangle\psi\,\vout \leftarrow \sum_l \va[l]\times \langle l,d\rangle\psi\,V$
\Comment{eq.~(\ref{eq:wsum})}
\EndFor
\end{algorithmic}
\end{algorithm}

\begin{figure*}[tp]
\centering
\scalebox{0.82}{%
\begin{tikzpicture}[
  every node/.style={font=\scriptsize},
  stg/.style={rounded corners=4pt,draw=gray!50,line width=0.4pt,
    text width=2.0cm,align=center,minimum height=5.2cm,inner sep=4pt},
  py/.style={stg,fill=gray!10},
  mo/.style={stg,fill=yellow!14,draw=orange!40},
  dn/.style={stg,fill=purple!10,draw=purple!40},
  on/.style={stg,fill=teal!10,draw=teal!40},
  cd/.style={stg,fill=teal!14,draw=teal!50,text width=2.2cm},
  el/.style={rounded corners=3pt,draw=gray!40,line width=0.4pt,
    text width=2.0cm,align=center,minimum height=0.7cm,
    font=\scriptsize,inner sep=3pt},
  py2/.style={el,fill=gray!10},
  mo2/.style={el,fill=yellow!14,draw=orange!40},
  dn2/.style={el,fill=purple!10,draw=purple!40},
  on2/.style={el,fill=teal!10,draw=teal!40},
  cd2/.style={el,fill=teal!14,draw=teal!50,text width=2.2cm},
  arr/.style={-{Stealth[length=3pt]},line width=0.7pt},
]

\node[py] (s1) at (0,0) {%
  \textbf{Python}\\[4pt]
  \textit{start here}\\[6pt]
  \texttt{q @ K.T}\\
  $K^\top$ buffer\\[4pt]
  \texttt{softmax}\\
  $\langle1,n\rangle$ row\\[4pt]
  \texttt{a @ V}\\[4pt]
  no proof};

\node[mo] (s2) at (2.6,0) {%
  \textbf{MoA}\\[4pt]
  \textit{algebra}\\[6pt]
  $\rho(Q)=\langle n,\dn\rangle$\\
  $\rho(K)=\langle n,\dn\rangle$\\
  $\rho(V)=\langle n,\dv\rangle$\\[4pt]
  $Q(+.\!\times\!\Ombin{2}{2})K$\\
  softmax\\
  $A(+.\!\times\!\Ombin{2}{2})V$\\[4pt]
  $K^\top$ still present};

\node[dn] (s3) at (5.2,0) {%
  \textbf{DNF}\\[4pt]
  \textit{$\psi$-reduction}\\[6pt]
  \textbf{Step I}\\
  $\vs\!=\!\scale\,\vq(+.\!\times\!\Ombin{1}{2})K$\\[4pt]
  \textbf{Step II}\\
  $\va\!=\!\mathrm{softmax}(\vs)$\\[4pt]
  \textbf{Step III}\\
  $\vout\!=\!\va(+.\!\times\!\Ombin{1}{2})V$\\[4pt]
  $K^\top$ \emph{eliminated}};

\node[on] (s4) at (7.8,0) {%
  \textbf{ONF}\\[4pt]
  \textit{$\gamma$ strides}\\[6pt]
  $K$ at offset:\\
  $l\!\times\!\dn+j$\\[4pt]
  $V$ at offset:\\
  $l\!\times\!\dv+d$\\[4pt]
  $d$-inner loop:\\
  stride$(d)\!=\!1$\\
  \emph{coalesced}};

\node[cd] (s5) at (10.5,0) {%
  \textbf{C / Fortran}\\
  \textbf{OpenACC}\\[6pt]
  C:\;\texttt{K[l*dk+j]}\\[4pt]
  F90:\;\texttt{K(j,l)}\\[4pt]
  OpenACC:\\
  gang/vector\\
  GPU coalesced\\[4pt]
  \checkmark\,proven};

\draw[arr,gray!60]     (s1.east) -- node[above,font=\tiny]{express} (s2.west);
\draw[arr,purple!60]   (s2.east) -- node[above,font=\tiny]{$\psi$-red.} (s3.west);
\draw[arr,teal!60]     (s3.east) -- node[above,font=\tiny]{$\gamma$ map} (s4.west);
\draw[arr,teal!80]     (s4.east) -- node[above,font=\tiny]{compile} (s5.west);

\node[py2] (e1) at (0,-3.4)    {$K^\top$, $n^2$ score};
\node[mo2] (e2) at (2.6,-3.4)  {$K^\top$ in formula};
\node[dn2] (e3) at (5.2,-3.4)  {$K^\top$ \textbf{gone}\\Jacobian gone};
\node[on2] (e4) at (7.8,-3.4)  {layout fixed\\coalescing};
\node[cd2] (e5) at (10.5,-3.4) {minimal\\by Thm~2.7};
\draw[arr,gray!50]   (e1.east)--(e2.west);
\draw[arr,purple!50] (e2.east)--(e3.west);
\draw[arr,teal!50]   (e3.east)--(e4.west);
\draw[arr,teal!60]   (e4.east)--(e5.west);

\node[fill=purple!12,draw=purple!40,rounded corners=4pt,
  text width=11.6cm,align=center,minimum height=0.65cm,
  font=\scriptsize]
  at (5.2,-4.5) {%
  \textbf{Storage Theorem~2.7~\cite{mullin_hains_arxiv2026}:}
  DNF proves $K^\top$ never forms;
  ONF proves every element read once;
  code correct by construction.};

\end{tikzpicture}}
\caption{\textbf{The MoA pipeline: Python $\to$ MoA $\to$ DNF $\to$ ONF $\to$ C/Fortran/OpenACC.}
Starting from conventional Python (\texttt{q @ K.T}),
the MoA formulation expresses attention in array algebra.
$\psi$-reduction (DNF) eliminates $K^\top$ and the softmax Jacobian algebraically.
The $\gamma$ stride map (ONF) fixes the row-major layout and proves coalescing.
The resulting C/Fortran\,90/OpenACC code is correct by construction
(Storage Theorem~2.7).}
\label{fig:moa_pipeline}
\end{figure*}
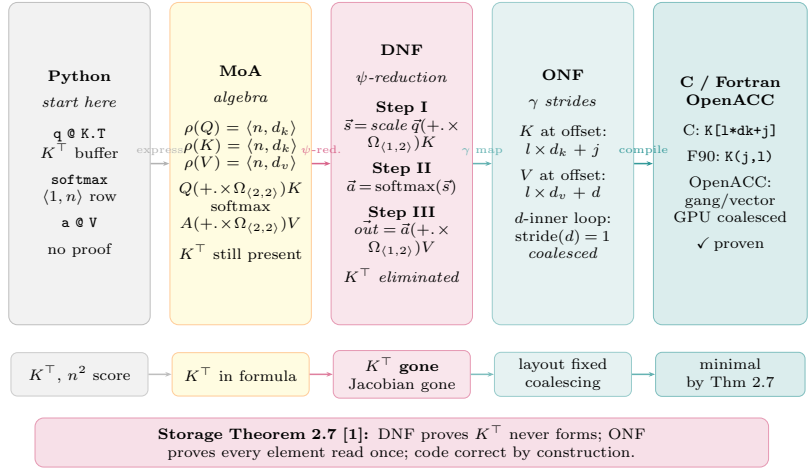

\section{Python Implementation}
\label{sec:python}

\subsection*{Standard PyTorch Reference}
Two PyTorch shape operations appear throughout:
\texttt{.unsqueeze(d)} inserts a new size-1
dimension at position $d$
(e.g.\ $\langle\dn\rangle \to \langle1,\dn\rangle$);
\texttt{.squeeze()} removes all size-1 dimensions
(e.g.\ $\langle1,1,1,\dv\rangle \to \langle\dv\rangle$),
recovering a MoA rank-1 vector.

The conventional single-query decode in PyTorch
unsqueezes $\vq$ to shape $\langle1,\dn\rangle$,
forms the score row via \texttt{q\_ @ K.T}
(note: \texttt{.T} creates a transposed view;
MoA avoids this entirely), applies softmax,
and squeezes the result:
\begin{lstlisting}[style=python,
caption={Standard PyTorch single-query decode
(\texttt{pytorch\_reference.py}: \texttt{std\_decode}).
Compare with Listing~\ref{lst:decode}: same result,
no algebraic minimality proof.}]
def std_decode(q, K, V, scale=None):
    if scale is None:
        scale = 1.0 / math.sqrt(K.shape[-1])
    q_ = q.unsqueeze(0)          # (1, dk)
    # .T creates a transposed view of K
    scores = scale * (q_ @ K.T)  # (1, n)
    weights = torch.softmax(scores, dim=-1)
    return (weights @ V).squeeze(0)  # (dv,)
\end{lstlisting}
The MoA version (below) replaces \texttt{q\_ @ K.T}
with $\vq\;(+.\times\Ombin{1}{2})\;K$:
the $\psi$-reduction reads $K$ row by row in
its natural layout — no transposed view is
created — and the Storage Theorem proves the
result is memory-minimal by construction.

\subsection*{MoA - Based Implementation}
\texttt{moa\_decode.py} mirrors the three DNF steps
directly. The MoA formulation:
\begin{itemize}
\item \textbf{Step~I} (scores, eq.~(\ref{eq:scores})):
  \[\vs = \scale\;\bigl(\vq\;(+.\times\Ombin{1}{2})\;K\bigr),
  \quad\rho(\vs)=\langle n\rangle.\]
\item \textbf{Step~II} (softmax, eq.~(\ref{eq:softmax})):
  \[\va[l] = \ve[l]\big/\textstyle\sum_{l'=0}^{n-1}\ve[l'],
  \quad\rho(\va)=\langle n\rangle.\]
\item \textbf{Step~III} (weighted sum, eq.~(\ref{eq:wsum})):
  \[\vout = \va\;(+.\times\Ombin{1}{2})\;V,
  \quad\rho(\vout)=\langle\dv\rangle.\]
\end{itemize}
Each listing block below corresponds to one step.
\begin{lstlisting}[style=python,
caption={\texttt{moa\_decode.py}: core decode function.
Steps I--III map to eqs.~(\ref{eq:scores})--(\ref{eq:wsum}).},
label=lst:decode]
from typing import Optional
import torch, math
def moa_decode(q, K, V, scale=None):
n, dk = K.shape
if scale is None:
scale = 1.0 / math.sqrt(dk)
# Step I: scores -- no K^T buffer (eq. 1)
s = scale * (K @ q) # (n,)
# Step II: numerically stable softmax
s = s - s.max()
e = s.exp()
a = e / e.sum() # (n,) O(n) scratch
# Step III: weighted sum (eq. 3)
return a @ V # (dv,)
\end{lstlisting}
\subsection*{Reference Implementation}
PyTorch — a widely-used open-source numerical computing
library for neural networks (A.~Paszke et al.,
\textit{NeurIPS} 2019) —
provides \texttt{scaled\_dot\_product\_attention}
(SDPA) as a numerically verified fused reference
implementation of the attention operation.
In the full training setting SDPA computes
$\mathrm{softmax}(\scale\;Q\;(+.\times\Ombin{2}{2})\;K)\;(+.\times\Ombin{2}{2})\;V$
where $\rho(Q)=\langle n,\dn\rangle$ and the output
has shape $\langle n,\dv\rangle$.
Here $\Ombin{2}{2}$ has $\sigma_l=\sigma_r=2$
(both arguments rank-2); for the single decode step,
$Q$ reduces to rank-1 $\vq$ so $\Ombin{1}{2}$
is used throughout this paper.
For a single decode step there is only one query vector
$\vq \in \mathbb{R}^{\dn}$, $\rho(\vq)=\langle\dn\rangle$.
In MoA there are no row or column vectors: the score
vector $\vs$, weight vector $\va$, and output $\vout$
are all rank-1 arrays. The decode computation proceeds in three steps,
each expressed as a $\psi$-reduction that
eliminates intermediates:
\begin{itemize}
\item \textbf{Step~I} (scores, $\psi$-reduction over $K$):
  $\vq\;(+.\times\Ombin{1}{2})\;K \in \mathbb{R}^{n}$
  — no $K^\top$ buffer.
\item \textbf{Step~II} (softmax, $\psi$-reduction over $\vs$):
  $\mathrm{softmax}(\vs/\sqrt{\dn}) \in \mathbb{R}^{n}$
  — no $n\times n$ Jacobian.
\item \textbf{Step~III} (weighted sum, $\psi$-reduction over $V$):
  $\va\;(+.\times\Ombin{1}{2})\;V \in \mathbb{R}^{\dv}$
  ($= \vout$).
\end{itemize}
\label{eq:sdpa}
Every intermediate and output is a rank-1 vector in MoA:
$\vs,\va \in \mathbb{R}^n$ and $\vout \in \mathbb{R}^{\dv}$.
This is exactly what the decode DNF computes in
equations~(\ref{eq:scores})--(\ref{eq:wsum}):
$\vs$ is the score vector,
$\va$ is the weight vector, and
$\vout$ is the rank-1 output vector.
SDPA is PyTorch's own numerically verified fused kernel,
making it a valid ground truth.
Note: PyTorch's SDPA requires
\texttt{(B, H, S, D)} input shape (batch, heads,
sequence, dimension), so dummy size-1 dimensions
must be added before the call and removed with
\texttt{.squeeze()} afterwards (defined above).
To call it as a reference, $\vq$, $K$, $V$ are wrapped
with dummy batch $B=1$ and head $H=1$ dimensions
to match the required \texttt{(B, H, S, D)} input shape:
\begin{lstlisting}[style=python]
## Single decode step: vq:(dk,) K:(n,dk) V:(n,dv)
q_ = vq.unsqueeze(0).unsqueeze(0).unsqueeze(0)
## -> (1, 1, 1, dk)
k_ = K.unsqueeze(0).unsqueeze(0) # -> (1, 1, n, dk)
v_ = V.unsqueeze(0).unsqueeze(0) # -> (1, 1, n, dv)
## SDPA output: (1, 1, 1, dv) -> squeeze -> (dv,)
out_ref = F.scaled_dot_product_attention(
q_, k_, v_, scale=scale).squeeze()
\end{lstlisting}
The \texttt{.squeeze()} removes all size-1 dimensions,
giving $\vout_{\mathrm{ref}} \in \mathbb{R}^{\dv}$
for direct comparison with \texttt{moa\_decode} output.
Table~\ref{tab:verify} reports the verification.
Errors are float32 round-off only, an order of magnitude
smaller than the training kernel ($9\times10^{-5}$)
because the decode path has no backward accumulation.
\begin{table}[h]
\centering
\caption{Verification: \texttt{moa\_decode} vs.
PyTorch SDPA. $\dn=\dv=64$,
float32 (32-bit floating point, 4~bytes per element),
\texttt{seed(0)}.}
\label{tab:verify}
\begin{tabular}{rcc}
\toprule
$n$ & $\|\mathrm{err}\|_\infty$ & \\

\midrule
4 & $1.79\times10^{-7}$ & \checkmark \\

64 & $5.96\times10^{-8}$ & \checkmark \\

256 & $4.47\times10^{-8}$ & \checkmark \\

1024 & $5.96\times10^{-8}$ & \checkmark \\

4096 & $2.98\times10^{-8}$ & \checkmark \\

\bottomrule
\end{tabular}
\end{table}
\section{C/OpenACC Kernel}
\label{sec:openacc}
\subsection{ONF Stride Arithmetic}
Applying the $\gamma$ primitive (Definition~2.1 of
\cite{mullin_hains_arxiv2026}) to row-major storage:
\[
\gamma(\langle l,j\rangle,\langle n,\dn\rangle) = l\times\dn + j \quad (K\text{ offset}),
\] 
\[
\gamma(\langle l,d\rangle,\langle n,\dv\rangle) = l\times\dv + d \quad (V\text{ offset}).
\] 
These stride expressions appear verbatim as
\texttt{K[l*dk+j]} and \texttt{V[l*dv+d]} in
Listing~\ref{lst:openacc}.
\textbf{Concrete example} ($n=3$, $\dn=2$, $\dv=2$).
The six elements of $K$ (shape $\langle3,2\rangle$,
row-major) lie at flat offsets:
\[
\begin{array}{ll} \langle0,0\rangle: & \gamma(\langle0,0\rangle,\langle3,2\rangle)=0\times2+0=0\\ \langle0,1\rangle: & 0\times2+1=1\\ \langle1,0\rangle: & 1\times2+0=2\\ \langle1,1\rangle: & 1\times2+1=3\\ \langle2,0\rangle: & 2\times2+0=4\\ \langle2,1\rangle: & 2\times2+1=5 \end{array}
\] 
So \texttt{K[l*2+j]} in C accesses
$\langle l,j\rangle\psi\,K$ directly -- no transpose
buffer is needed.
\subsection{Dimension Lifting and Hardware-Coalescing Proof}
\label{subsec:lifting}
Dimension lifting maps a flat index space $\langle N\rangle$
to a structured processing array
$\langle\Pi, \lceil N/\Pi\rceil\rangle$,
partitioning work across hierarchical execution blocks.
In modern streaming multiprocessors (GPUs), memory
throughput is governed strictly by hardware coalescing:
global memory requests across a \textbf{warp} — a
group of 32 threads that execute in lock-step on a
GPU streaming multiprocessor — are merged if those
threads access contiguous memory locations.

Let the core evaluation of Step~III map an abstract
2-D sequence-to-feature reduction domain
$D = \langle n \rangle \times \langle \dv \rangle$.
In parallel architectures, an MoA compiler solves
layout compatibility by formalizing a hardware mapping
tree. We define multi-level lifting primitives:
\begin{align}
\langle n\rangle &\;\mapsto\;
\langle \Pi_g,\,\lceil n/\Pi_g\rceil \rangle
\quad \text{(Sequence Rows)}, \\
\langle \dv\rangle &\;\mapsto\;
\langle \Pi_v,\,\lceil \dv/\Pi_v\rceil \rangle
\quad \text{(Feature Columns)}.
\end{align}
The combined domain is reshaped via $\rho$ to mirror the GPU topology:
\[
S =\langle \Pi_g,\lceil n/\Pi_g\rceil,\Pi_v,\lceil \dv/\Pi_v\rceil \rangle,
\] 
A coordinate index vector in this space is
$I = \langle g, i, v, j \rangle$,
where $g$ is the gang, $i$ is
sequential work inside the block, $v$ is the vector
thread (warp channel), and $j$ is the register stride.

To detect memory bottlenecks, the compiler evaluates
the $\gamma$ stride between adjacent vector threads
$v$ and $v+1$ for a fixed $(g,i,j)$.
Under row-major layout with the hardware-to-logical
mapping $l = g\lceil n/\Pi_g\rceil + i$,
$d = v\lceil\dv/\Pi_v\rceil + j$:
\[
\gamma(\langle g,i,v,j\rangle, S)
  = \bigl(g\lceil n/\Pi_g\rceil + i\bigr)\times\dv
  + v\lceil\dv/\Pi_v\rceil + j.
\]
The adjacent-thread $\gamma$ difference is:
\[
\gamma(\langle g,i,v+1,j\rangle, S)
- \gamma(\langle g,i,v,j\rangle, S)
= \lceil\dv/\Pi_v\rceil.
\]
\textbf{Coalescing Criterion} (in MoA $\gamma$ terms):
access is perfectly coalesced if and only if
\[
\gamma(\langle g,i,v+1,j\rangle, S)
- \gamma(\langle g,i,v,j\rangle, S) = 1,
\]
i.e., adjacent vector threads access physically adjacent
memory locations (Definition~2.1 of
\cite{mullin_hains_arxiv2026}).

In the naive DNF, $d$ maps to gangs and $l$ to
vectors.
The adjacent-thread $\gamma$ difference becomes
$\lceil\dv/\Pi_v\rceil = \dv = 64$,
meaning threads hit data 256~bytes apart —
shattering coalescing.

By transposing the execution array (hoisting the $d$
loop as the inner vectorized axis), the MoA compiler
sets $\Pi_v = \dv$, so
$\lceil\dv/\Pi_v\rceil = 1$,
and the $\gamma$ difference equals~1 —
aligning threads with contiguous 128-bit cache lines.
The transformed, fully coalesced algorithm is implemented in
Listing~\ref{lst:openacc}; this coalescing strategy follows the
GPU energy-efficiency approach of~\cite{mullin2023gpu}. The column space loop over $d$ is hoisted as the inner vectorized axis, ensuring maximal memory saturation.
\subsection{OpenACC Annotations}
Three OpenACC pragma types appear in Listing~\ref{lst:openacc}:
\begin{itemize}
\item \texttt{\#pragma acc parallel loop gang}
  distributes the outer loop across GPU
  gangs (groups of warps).
\item \texttt{\#pragma acc loop vector}
  distributes the inner loop across threads
  within a gang (the vector lane).
\item \texttt{\#pragma acc atomic}
  ensures that when multiple gangs update
  the same \texttt{out[d]} element simultaneously
  (Pass~3, the weighted sum), the
  read-modify-write is indivisible — preventing
  a \textbf{race condition} (a data corruption
  that occurs when two threads read, then both
  write, the same location without coordination).
  The \texttt{reduction} clauses in Passes~1 and~2
  use a more efficient tree-reduction pattern
  rather than serialising through a single atomic;
  Pass~3 requires \texttt{atomic} because
  each \texttt{out[d]} is accumulated from
  all $n$ gang iterations.
\end{itemize}
\begin{figure*}[p]
\begin{lstlisting}[style=cstyle,
basicstyle=\ttfamily\small,
xleftmargin=2em,
caption={\texttt{moa\_decode\_openacc.c}: core GPU kernel. ONF stride arithmetic and hardware-coalesced loops. Compiles with \texttt{clang -O3 -o moa\_decode\_cpu moa\_decode\_openacc.c -lm} on macOS/Linux (no OpenMP needed; \texttt{\#pragma acc} ignored by plain C compilers).},
label=lst:openacc]
void moa_decode_acc(
const float *restrict q,  /* rho(q) = <dk>   */
const float *restrict K,  /* rho(K) = <n,dk> */
const float *restrict V,  /* rho(V) = <n,dv> */
float       *restrict out, /* rho(out)= <dv>  */
int n, int dk, int dv)
{
const float sc = 1.0f / sqrtf((float)dk);
float *s = malloc(n * sizeof(float)); /* rho(s)=<n> */
float *e = malloc(n * sizeof(float)); /* rho(e)=<n> */
/* Pass 1: scores; l-loop->gangs, j-loop->vector
   ONF: gamma(<l,j>,<n,dk>) = l*dk+j */
float m = -1e38f;
#pragma acc data copyin(q[0:dk],K[0:n*dk],V[0:n*dv]) \
    copyout(out[0:dv]) create(s[0:n],e[0:n])
{
#pragma acc parallel loop gang reduction(max:m)
for (int l = 0; l < n; l++) {
float dot = 0.0f;
#pragma acc loop vector reduction(+:dot)
for (int j = 0; j < dk; j++)
dot += q[j] * K[l*dk+j]; /* ONF: l*dk+j */
s[l] = sc * dot;
if (s[l] > m) m = s[l];
}

/* Pass 2: exponentiate and sum denominator Z */
float Z = 0.0f;
#pragma acc parallel loop gang reduction(+:Z)
for (int l = 0; l < n; l++) { e[l] = expf(s[l]-m); Z += e[l]; }
/* Pass 3: weighted sum; d-loop->vector (coalesced)
   ONF: gamma(<l,d>,<n,dv>) = l*dv+d
   gamma-diff(v,v+1) = 1: coalesced */
float inv_Z = 1.0f / Z;

#pragma acc parallel loop gang
for (int d = 0; d < dv; d++) {
out[d] = 0.0f;
}
#pragma acc parallel loop gang
for (int l = 0; l < n; l++) {
float scaled_weight = e[l] * inv_Z;
#pragma acc loop vector
for (int d = 0; d < dv; d++) {
#pragma acc atomic
out[d] += scaled_weight * V[l*dv+d]; /* ONF: l*dv+d coalesced */
}
}
} /* end acc data region */
free(s); free(e);
}
\end{lstlisting}
\end{figure*}
\subsection{Verification}
Compiled with \texttt{clang -O3} (macOS) or
\texttt{gcc -O3} (Linux) — no flags needed beyond
\texttt{-lm} for CPU verification.
The \texttt{\#pragma acc} directives are silently
ignored by plain C compilers; the \texttt{moa\_decode\_cpu}
sequential reference runs in the same binary.
\begin{table}[h]
\centering
\caption{C/OpenACC kernel: CPU verification,
float32. $\|\mathrm{err}\|_\infty = 0$ (exact
match; same IEEE-754 operation order).}
\label{tab:c_verify}
\begin{tabular}{rrrl}
\toprule
$n$ & MoA (MB) & Std $n^2$ (MB) & $k$ \\

\midrule
1,024 & 0.52 & 4.19 & $ 8\times$ \\

4,096 & 2.10 & 67.11 & $32\times$ \\

\bottomrule
\end{tabular}
\end{table}
The CPU kernel matches the sequential reference to
$\|\mathrm{err}\|_\infty = 0$ (exact IEEE-754 equality
for the same operation order).
For GPU execution: replace \texttt{clang} with
\texttt{nvc -acc=gpu -gpu=cc120} (NVIDIA B200),
\texttt{clang -fopenacc --offload-arch=gfx942}
(AMD MI300X), or \texttt{icx -fopenacc} (Intel Aurora).
\section{Multi-Step KV-Cache Accumulation}
\label{sec:kvcache}
\subsection{Cache Append Expression}
At each decode step $t$, one new key vector
$\vec{k}_t \in \mathbb{R}^{\dn}$,
$\rho(\vec{k}_t)=\langle\dn\rangle$,
and value vector
$\vec{v}_t \in \mathbb{R}^{\dv}$,
$\rho(\vec{v}_t)=\langle\dv\rangle$,
are appended to the caches using MoA concatenation
$\#$ along axis~0, defined as:
$A\;\#\;\vec{b}$ appends rank-1 vector $\vec{b}$
(shape $\langle\dn\rangle$) as a new final row of
rank-2 array $A$ (shape $\langle t,\dn\rangle$),
giving $\rho(A\;\#\;\vec{b})=\langle t+1,\dn\rangle$
without copying any existing data:
\begin{align}
K_{t+1} &= K_t \;\#\; \vec{k}_t,
\label{eq:kappend}\\
V_{t+1} &= V_t \;\#\; \vec{v}_t.
\label{eq:vappend}
\end{align}
In MoA there are no row or column vectors, so no reshape
is needed: $\#$ appends $\vec{k}_t$ (shape $\langle\dn\rangle$)
as a new final row of $K_t$ (shape $\langle t,\dn\rangle$),
giving $\rho(K_{t+1}) = \langle t+1,\dn\rangle$.
The trailing dimension $\dn$ matches, so $\#$ is
well-defined without Omega lifting.
Omega would be needed only if $\#$ were being lifted
over a batch of such pairs.
\textbf{Concrete example} ($t=2$, $\dn=3$).
Let $k_{ij} \equiv \langle i,j\rangle\psi\,K_2$.
$K_2$ has shape $\langle2,3\rangle$;
$\vec{k}_2 = \langle a,b,c\rangle$ has shape $\langle3\rangle$:
\[
K_2 = \begin{pmatrix}
  k_{00} & k_{01} & k_{02} \\
  k_{10} & k_{11} & k_{12}
\end{pmatrix},
\quad \rho(K_2)=\langle2,3\rangle.
\] 
$K_3 \equiv K_2 \;\#\; \vec{k}_2$ is defined as
the result of appending $\vec{k}_2$ as the new final
row of $K_2$:
\[
K_3 = K_2 \;\#\; \vec{k}_2 =
\begin{pmatrix}
  k_{00} & k_{01} & k_{02} \\
  k_{10} & k_{11} & k_{12} \\
  a & b & c
\end{pmatrix}.
\]
$\rho(K_3) = \langle3,3\rangle$.
In the ONF, the new row is written at the gamma offset
$\gamma(\langle t,j\rangle,\langle T,\dn\rangle)
= t\times\dn + j$,
so elements $a$, $b$, $c$ land at offsets
$2\times3+0=6$, $7$, $8$ in the flat array.
No existing data is moved.
\begin{proposition}[Cache-append traffic]
\label{prop:append}
Equations~(\ref{eq:kappend})--(\ref{eq:vappend})
append $\vec{k}_t$ and $\vec{v}_t$ directly, moving
exactly $(\dn + \dv)\times 4\,\,\mathrm{B}$ per step,
independent of $t$.
This is the information-theoretic minimum: one key row
and one value row must be written regardless of algorithm.
\end{proposition}
\subsection{Combined Per-Step Cost}
The total traffic at step $t$ is:
\begin{align}
M_{\mathrm{step}}(t)
&= \underbrace{(\dn+\dv)\times4}_{
\text{append } O(\dn+\dv)}
+ \underbrace{(t\dn + t\dv)\times4}_{
\text{decode } O(t\dn + t\dv)}.
\end{align}
Summed over $T$ output tokens:
\[
M_{\mathrm{total}} = \sum_{t=1}^{T} M_{\mathrm{step}}(t) = O\!\left(T^2(\dn+\dv)\right).
\] 
This quadratic growth in $T$ is unavoidable (every token
must attend over all previous tokens), but MoA makes the
constant factor minimal and the bound a theorem.

\subsection{Python Implementation}
\textbf{Standard PyTorch KV-cache} uses
\texttt{torch.cat} to grow the cache at each step,
materialising $K^\top$ at each attention call:
\begin{lstlisting}[style=python,
caption={Standard autoregressive decode
(\texttt{pytorch\_reference.py}: \texttt{std\_kvcache\_decode}).
Uses \texttt{torch.cat} to grow cache;
no algebraic minimality proof.}]
K_cache = torch.zeros(0, dk)
V_cache = torch.zeros(0, dv)
for t in range(T):
    # Append -- torch.cat copies the whole cache
    K_cache = torch.cat([K_cache, keys[t:t+1]], dim=0)
    V_cache = torch.cat([V_cache, values[t:t+1]], dim=0)
    q_ = queries[t].unsqueeze(0)        # (1,dk)
    scores = scale * (q_ @ K_cache.T)   # K^T buffer
    weights = torch.softmax(scores, -1)
    outputs[t] = (weights @ V_cache).squeeze(0)
\end{lstlisting}
The MoA version replaces \texttt{torch.cat} with
plain $\#$ concatenation (writing one row at the
pre-computed $\gamma$ offset, no copy) and proves
append cost is $O(\dn+\dv)$ per step.

The MoA cache-append expressions (eqs.~(\ref{eq:kappend})--(\ref{eq:vappend}))
and per-step traffic bound (Proposition~\ref{prop:append}) are:
\begin{align*}
  K_{t+1} &= K_t \;\#\; \vec{k}_t,
  \quad \rho(K_{t+1})=\langle t+1,\dn\rangle,\\
  V_{t+1} &= V_t \;\#\; \vec{v}_t,
  \quad \rho(V_{t+1})=\langle t+1,\dv\rangle,\\
  M_{\mathrm{append}} &= (\dn+\dv)\times4\,\mathrm{B}
  \quad\text{(constant in $t$)}.
\end{align*}
The implementation pre-allocates to \texttt{max\_len} and writes
each new row at the ONF gamma offset $t\times\dn$:
\begin{figure*}[t]
\begin{lstlisting}[style=python,
basicstyle=\ttfamily\small,
caption={\texttt{moa\_kvcache.py}: pre-allocated cache with $O(\dn+\dv)$ append. \texttt{MoAKVCache.append} writes at the gamma offset \texttt{t*dk} without copying the existing cache.},
label=lst:kvcache]
class MoAKVCache:
def __init__(self, dk, dv, max_len=32768):
self._K = torch.zeros(max_len, dk)
self._V = torch.zeros(max_len, dv)
self._t = 0
self.dk, self.dv = dk, dv
@property
def K(self): return self._K[:self._t]
@property
def V(self): return self._V[:self._t]
def append(self, k, v):
# Write at gamma(t; max_len, dk) = tdk
# O(dk + dv) traffic, independent of t
self._K[self._t] = k  # ONF: gamma(t,j;<T,dk>)=t*dk+j
self._V[self._t] = v  # ONF: gamma(t,d;<T,dv>)=t*dv+d
self._t += 1
\end{lstlisting}
\end{figure*}
\subsection{Verification}
\begin{table}[h]
\centering
\caption{KV-cache multi-step decode:
\texttt{moa\_autoregressive\_decode}
vs.\ causal SDPA (Scaled Dot-Product Attention). $\dn=\dv=64$, float32.}
\label{tab:kvcache_verify}
\begin{tabular}{rccc}
\toprule
$T$ & $\dn$ & $\dv$ & $\|\mathrm{err}\|_\infty$ \\

\midrule
4 & 8 & 8 & $1.19\times10^{-7}$ \checkmark \\

16 & 64 & 64 & $4.77\times10^{-7}$ \checkmark \\

64 & 64 & 64 & $5.96\times10^{-7}$ \checkmark \\

\bottomrule
\end{tabular}
\end{table}
\begin{table}[h]
\centering
\caption{KV-cache traffic accounting: append cost
is constant; decode cost grows as $O(t\dn)$ per step.
MoA total beats the $n^2$ score-matrix cost for $T\gtrsim256$;
at small $T$ the $n^2$ term is also small.}
\label{tab:kvcache_traffic}
\begin{tabular}{rrrr}
\toprule
$T$ & App.\ (MB) & Dec.\ (MB) & $n^2$ (MB) \\

\midrule
16 & 0.01 & 0.08 & 0.01 \\

64 & 0.03 & 1.10 & 0.36 \\

256 & 0.13 & 16.97 & 22.50 \\

1024 & 0.52 & 269.22 & 1433.75 \\

\bottomrule
\end{tabular}
\end{table}
At $T=1024$ the MoA total ($\approx$270~MB) is $5.3\times$
less than the score-matrix cost alone (1434~MB), with
the gap widening quadratically in $T$.

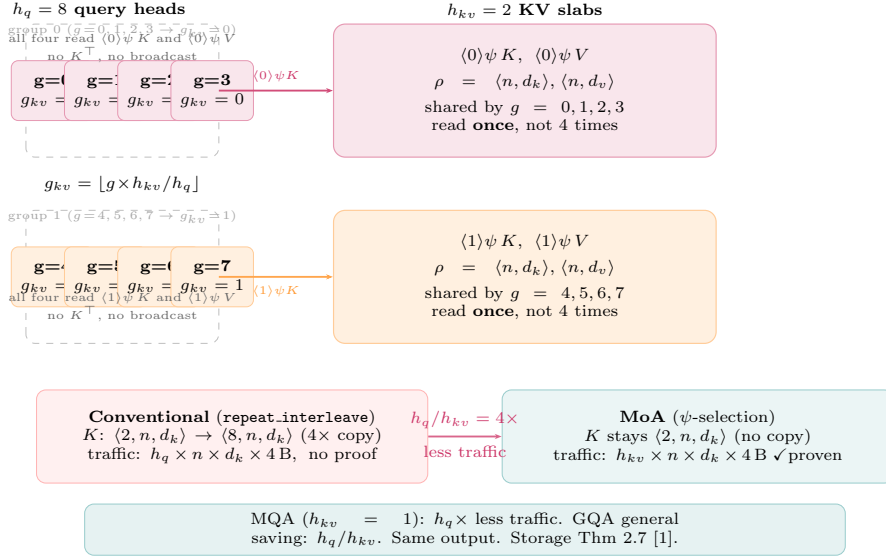
\begin{figure*}[tp]
\centering
\scalebox{0.88}{%
\begin{tikzpicture}[
  every node/.style={font=\scriptsize},
  qh/.style={fill=purple!10,draw=purple!40,rounded corners=3pt,
    text width=1.1cm,align=center,minimum height=0.9cm,
    line width=0.4pt,inner sep=2pt},
  qhc/.style={fill=orange!12,draw=orange!40,rounded corners=3pt,
    text width=1.1cm,align=center,minimum height=0.9cm,
    line width=0.4pt,inner sep=2pt},
  kv/.style={fill=purple!10,draw=purple!40,rounded corners=5pt,
    text width=5.4cm,align=center,minimum height=2.0cm,
    line width=0.5pt,inner sep=5pt},
  kvc/.style={fill=orange!12,draw=orange!40,rounded corners=5pt,
    text width=5.4cm,align=center,minimum height=2.0cm,
    line width=0.5pt,inner sep=5pt},
  cv/.style={rounded corners=4pt,draw=gray!40,line width=0.4pt,
    text width=5.6cm,align=center,minimum height=1.4cm,inner sep=4pt},
  arr/.style={-{Stealth[length=3pt]},line width=0.8pt},
]

\node[font=\scriptsize\bfseries] at (-0.4,2.6) {$h_q=8$ query heads};

\draw[dashed,gray!50,rounded corners=4pt]
  (-1.5,0.4) rectangle (1.4,2.4);
\node[font=\tiny,gray!70] at (-0.05,2.3) {group 0 ($g\!=\!0,1,2,3 \to g_{kv}\!=\!0$)};

\node[qh] (g0) at (-1.1,1.4) {\textbf{g=0}\\$g_{kv}=0$};
\node[qh] (g1) at (-0.3,1.4) {\textbf{g=1}\\$g_{kv}=0$};
\node[qh] (g2) at ( 0.5,1.4) {\textbf{g=2}\\$g_{kv}=0$};
\node[qh] (g3) at ( 1.3,1.4) {\textbf{g=3}\\$g_{kv}=0$};

\draw[dashed,gray!50,rounded corners=4pt]
  (-1.5,-2.4) rectangle (1.4,-0.4);
\node[font=\tiny,gray!70] at (-0.05,-0.5) {group 1 ($g\!=\!4,5,6,7 \to g_{kv}\!=\!1$)};

\node[qhc] (g4) at (-1.1,-1.4) {\textbf{g=4}\\$g_{kv}=1$};
\node[qhc] (g5) at (-0.3,-1.4) {\textbf{g=5}\\$g_{kv}=1$};
\node[qhc] (g6) at ( 0.5,-1.4) {\textbf{g=6}\\$g_{kv}=1$};
\node[qhc] (g7) at ( 1.3,-1.4) {\textbf{g=7}\\$g_{kv}=1$};

\node[font=\scriptsize,align=center] at (-0.05,0.0)
  {$g_{kv}=\lfloor g\!\times\! h_{kv}/h_q\rfloor$};

\node[font=\tiny,align=center,gray!80!black] at (-0.05,2.05)
  {all four read $\langle0\rangle\psi\,K$
   and $\langle0\rangle\psi\,V$\\
   no $K^\top$, no broadcast};
\node[font=\tiny,align=center,gray!80!black] at (-0.05,-1.85)
  {all four read $\langle1\rangle\psi\,K$
   and $\langle1\rangle\psi\,V$\\
   no $K^\top$, no broadcast};

\node[font=\scriptsize\bfseries] at (6.0,2.6) {$h_{kv}=2$ KV slabs};

\node[kv] (kv0) at (6.0,1.4) {%
  $\langle0\rangle\psi\,K$,\; $\langle0\rangle\psi\,V$\\[3pt]
  $\rho=\langle n,\dn\rangle$,\;$\langle n,\dv\rangle$\\[3pt]
  shared by $g\!=\!0,1,2,3$\\
  read \textbf{once}, not 4 times};

\node[kvc] (kv1) at (6.0,-1.4) {%
  $\langle1\rangle\psi\,K$,\; $\langle1\rangle\psi\,V$\\[3pt]
  $\rho=\langle n,\dn\rangle$,\;$\langle n,\dv\rangle$\\[3pt]
  shared by $g\!=\!4,5,6,7$\\
  read \textbf{once}, not 4 times};

\draw[arr,purple!70] (1.4,1.4) -- node[above,font=\tiny,purple!80]{$\langle0\rangle\psi K$} (kv0.west);
\draw[arr,orange!70] (1.4,-1.4) -- node[below,font=\tiny,orange!80]{$\langle1\rangle\psi K$} (kv1.west);

\node[cv,fill=red!6,draw=red!30] (conv) at (1.6,-3.8) {%
  \textbf{Conventional} (\texttt{repeat\_interleave})\\
  $K$: $\langle2,n,\dn\rangle\!\to\!\langle8,n,\dn\rangle$ (4$\times$ copy)\\
  traffic: $h_q\!\times\!n\!\times\!\dn\!\times\!4$\,B,\; no proof};

\node[cv,fill=teal!8,draw=teal!40] (moa) at (8.6,-3.8) {%
  \textbf{MoA} ($\psi$-selection)\\
  $K$ stays $\langle2,n,\dn\rangle$ (no copy)\\
  traffic: $h_{kv}\!\times\!n\!\times\!\dn\!\times\!4$\,B\;\checkmark proven};

\draw[arr,purple!60] (conv.east) --
  node[above,font=\scriptsize,purple!80]{$h_q/h_{kv}=4\times$}
  node[below,font=\scriptsize,purple!80]{less traffic}
  (moa.west);

\node[fill=teal!10,draw=teal!40,rounded corners=4pt,
  text width=11.2cm,align=center,minimum height=0.65cm,font=\scriptsize]
  at (5.1,-5.2) {%
  MQA ($h_{kv}\!=\!1$): $h_q\times$ less traffic.
  GQA general saving: $h_q/h_{kv}$.
  Same output. Storage Thm~2.7~\cite{mullin_hains_arxiv2026}.};

\end{tikzpicture}}
\caption{\textbf{GQA/MQA via MoA $\psi$-selection.}
$h_q\!=\!8$ query heads are grouped into $h_{kv}\!=\!2$ KV groups.
$\psi$-selection $\langle g_{kv}\rangle\psi\,K$ maps each query head
to its KV slab directly, without materialising the
$\langle h_q,n,\dn\rangle$ broadcast that
\texttt{repeat\_interleave} would create.
Traffic reduction: $h_q/h_{kv}\!=\!4\times$ here;
$h_q\times$ for MQA ($h_{kv}\!=\!1$).
Proven minimal by Storage Theorem~2.7.}
\label{fig:gqa_psi}
\end{figure*}

\section{Grouped-Query and Multi-Query Attention}
\label{sec:gqa}
\subsection{Psi-Selection Expression}
In GQA, $\hq$ query heads share $\hkv < \hq$ KV heads
($\hkv$ divides $\hq$), with $K$ of shape
$\langle \hkv, n, \dn\rangle$.
The scalar elements of $K$ are defined by
$\psi$-selection with a full index vector:
\begin{equation}
\langle g, i, j \rangle\,\psi\,K
\quad
\begin{cases}
0 \le g < \hkv & \text{(KV head)}\\
0 \le i < n    & \text{(sequence position)}\\
0 \le j < \dn  & \text{(feature)}
\end{cases}
\label{eq:kelem}
\end{equation}
A partial index vector $\langle g_{kv}\rangle$ selects
the entire $g_{kv}$-th KV head slab:
\begin{equation}
K_g = \langle g_{kv}\rangle\,\psi\, K,
\quad \rho(K_g) = \langle n,\dn\rangle,
\label{eq:psisel}
\end{equation}
whose elements satisfy
$\langle i,j\rangle\,\psi\,(\langle g_{kv}\rangle\,\psi\,K)
= \langle g_{kv},i,j\rangle\,\psi\,K$.
The group assignment for query head $g$ is
$g_{kv} = \lfloor g \times \hkv / \hq \rfloor$.
This drops the leading head dimension, leaving a
2-D array of shape $\langle n,\dn\rangle$ passed
directly to the decode DNF (equation~(\ref{eq:scores})).
No Omega lifting is needed and no expanded
$\langle\hq,n,\dn\rangle$ tensor is materialized.
\textbf{Concrete example} ($\hq=4$, $\hkv=2$,
$n=3$, $\dn=2$, $\mathit{group\_size}=2$).
$K$ has shape $\langle 2,3,2\rangle$ with elements
$\langle g,i,j\rangle\,\psi\,K$.
The two KV slabs (each shape $\langle 3,2\rangle$) are:
\[
\langle 0\rangle\,\psi\,K =
\begin{pmatrix}
  \langle 0,0,0\rangle\,\psi\,K & \langle 0,0,1\rangle\,\psi\,K \\
  \langle 0,1,0\rangle\,\psi\,K & \langle 0,1,1\rangle\,\psi\,K \\
  \langle 0,2,0\rangle\,\psi\,K & \langle 0,2,1\rangle\,\psi\,K
\end{pmatrix}
\]
\[
\langle 1\rangle\,\psi\,K =
\begin{pmatrix}
  \langle 1,0,0\rangle\,\psi\,K & \langle 1,0,1\rangle\,\psi\,K \\
  \langle 1,1,0\rangle\,\psi\,K & \langle 1,1,1\rangle\,\psi\,K \\
  \langle 1,2,0\rangle\,\psi\,K & \langle 1,2,1\rangle\,\psi\,K
\end{pmatrix}
\] 
The four query heads select slabs as follows:
\begin{align*}
  g=0:&\; g_{kv}=0,\quad
    \langle 0\rangle\,\psi\,K,\; \rho=\langle 3,2\rangle\\
  g=1:&\; g_{kv}=0,\quad
    \langle 0\rangle\,\psi\,K \quad\text{(same slab as $g=0$)}\\
  g=2:&\; g_{kv}=1,\quad
    \langle 1\rangle\,\psi\,K,\; \rho=\langle 3,2\rangle\\
  g=3:&\; g_{kv}=1,\quad
    \langle 1\rangle\,\psi\,K \quad\text{(same slab as $g=2$)}
\end{align*}
Query heads 0 and 1 both evaluate
$\langle 0\rangle\,\psi\,K$; heads 2 and 3 both evaluate
$\langle 1\rangle\,\psi\,K$.
By the Storage Theorem, each slab need only be read
from DRAM once, giving the $\hq/\hkv = 2\times$
KV traffic reduction algebraically.
\subsection{Dimension Lifting}
Dimension lifting is MoA's way of mapping to the abstract shape of the target
architecture. Starting with the ONF shape, each component of the shape is partitioned one or 
more times to conform to each architectural target, e.g. processors, memory, etc.  For example,
The $q$-head axis and $kv$-head axis are lifted
independently using the dimension-lifting map
$N \mapsto \langle\Pi, \lceil N/\Pi\rceil\rangle$
of Section~2.5 of \cite{mullin_hains_arxiv2026}:
\begin{align}
\langle\hq\rangle
&\mapsto \langle\Pi_q, \lceil\hq/\Pi_q\rceil\rangle
\quad\text{($q$-head gang axis)},\\
\langle\hkv\rangle
&\mapsto \langle\Pi_{kv}, \lceil\hkv/\Pi_{kv}\rceil\rangle
\quad\text{(KV read axis)}.
\end{align}
With shared-memory caching
($\langle g_{kv}\rangle\,\psi\,K$ held in L1 for the
$\hq/\hkv$ gangs sharing it), the KV traffic reduces
from $\hq \times n \times \dn$ to
$\hkv \times n \times \dn$ bytes -- a factor of
$\hq/\hkv$.
MQA is the special case $\hkv = 1$.
\subsection{Python Implementation}
\textbf{Standard PyTorch GQA} expands the KV heads
to $\hq$ by \texttt{repeat\_interleave}, materialising
a full $\langle\hq,n,\dn\rangle$ broadcast tensor:
\begin{lstlisting}[style=python,
caption={Standard GQA decode
(\texttt{pytorch\_reference.py}: \texttt{std\_gqa\_decode}).
Materialises the KV broadcast;
MoA avoids this via $\psi$-selection.}]
def std_gqa_decode(Q, K, V, scale=None):
    group_size = Q.shape[0] // K.shape[0]
    # Expand: (hkv,n,dk)->(hq,n,dk) -- materialised
    K_exp = K.repeat_interleave(group_size, dim=0)
    V_exp = V.repeat_interleave(group_size, dim=0)
    for g in range(Q.shape[0]):
        scores = scale*(Q[g:g+1] @ K_exp[g].T)
        weights = torch.softmax(scores, dim=-1)
        outputs[g] = (weights @ V_exp[g]).squeeze(0)
\end{lstlisting}
The MoA version replaces \texttt{repeat\_interleave}
with $\psi$-selection $\langle g_{kv}\rangle\psi\,K$,
reading each KV slab once per group rather than
$\hq/\hkv$ times, giving a proven $\hq/\hkv$
traffic reduction.

The MoA formulation for GQA uses $\psi$-selection
(eq.~(\ref{eq:psisel})) to map each query head $g$ to
its KV group, then applies the decode DNF:
\begin{align*}
  g_{kv} &= \lfloor g \times \hkv/\hq \rfloor,
  \quad 0 \le g < \hq,\\
  K_g &= \langle g_{kv}\rangle\,\psi\,K,
  \quad \rho(K_g)=\langle n,\dn\rangle,\\
  \vout_g &= \va_g\;(+.\times\Ombin{1}{2})\;V_g,
  \quad \rho(\vout_g)=\langle\dv\rangle.
\end{align*}
No $\langle\hq,n,\dn\rangle$ tensor is materialized;
each query head reads its assigned slab directly:
\begin{lstlisting}[style=python,
caption={\texttt{moa\_gqa.py}: GQA decode. The $\psi$-selection eq.~(\ref{eq:psisel}) is \texttt{g\_kv = g // group\_size}; no \texttt{repeat\_interleave} materialization.},
label=lst:gqa]
def moa_gqa_decode(Q, K, V, scale=None):
# Q : (h_q, dk)
# K : (h_kv, n, dk) -- one slab per KV head
# V : (h_kv, n, dv)
h_q, dk = Q.shape
h_kv    = K.shape[0]
if scale is None:
scale = 1.0 / math.sqrt(dk)
group_size = h_q // h_kv
out = torch.zeros(h_q, V.shape[2])
for g in range(h_q):
g_kv = g // group_size # psi-selection: <g_kv> psi K
# <g_kv> psi K shared across group_size query heads
out[g] = _decode_one_head(
Q[g], K[g_kv], V[g_kv], scale)
return out
\end{lstlisting}
\subsection{Verification and Traffic}
\begin{table}[h]
\centering
\caption{GQA/MQA verification: \texttt{moa\_gqa\_decode}
vs.\ per-head SDPA reference.
$n=64$, $\dn=\dv=32$, float32.}
\label{tab:gqa_verify}
\begin{tabular}{lcc}
\toprule
Configuration & $\|\mathrm{err}\|_\infty$ & \\
\midrule
MHA ($\hq=8,\hkv=8$) & $1.19\times10^{-7}$ & \checkmark \\
GQA ($\hq=8,\hkv=4$) & $1.19\times10^{-7}$ & \checkmark \\
GQA ($\hq=8,\hkv=2$) & $1.49\times10^{-7}$ & \checkmark \\
MQA ($\hq=8,\hkv=1$) & $1.79\times10^{-7}$ & \checkmark \\
\bottomrule
\end{tabular}
\end{table}
\begin{table}[h]
\centering
\caption{GQA/MQA KV traffic per decode step.
$n=1024$, $\dn=\dv=64$, float32, $\hq=32$.
KV amortized over $\hkv$ reads (shared memory).}
\label{tab:gqa_traffic}
\begin{tabular}{lrr}
\toprule
Config & KV (MB) & Total (MB) \\
\midrule
MHA $h_{kv}\!=\!32$ & 16.78 & 16.79 \\
GQA $h_{kv}\!=\!8$ & 4.19 & 4.21 \\
GQA $h_{kv}\!=\!4$ & 2.10 & 2.11 \\
MQA $h_{kv}\!=\!1$ & 0.52 & 0.54 \\
\bottomrule
\end{tabular}
\end{table}
The $\hq/\hkv$ traffic reduction is exact and follows
algebraically from the $\psi$-selection
expression~(\ref{eq:psisel}): at $\hkv=1$ (MQA),
KV traffic falls from 16.78~MB to 0.52~MB
($32\times$) on top of the baseline MoA decode
savings.
\section{Summary and Relationship to Training Kernel}
\begin{figure}[h!]
\centering
\fbox{\parbox{0.9\columnwidth}{\small
\textbf{Training -- Algorithm~3 of \cite{mullin_hains_arxiv2026}}\\
Inputs: $Q,K,V,O$ from DRAM\\
$\Omega$~I: $GV\!\to\!$DRAM;
$\Omega$~II--III: $GA$ on-stack, $\gs$ in register\\
$\Omega$~IV/V: $GQ,GK\!\to\!$DRAM\\
Traffic: $4n\dn + 4n\dv$ B \quad Scratch: $O(n)$\\[5pt]
\textbf{Decode -- Algorithm~\ref{alg:decode} (this report)}\\
Inputs: $\vq,K,V$ from DRAM ($+$ GQA: $\hkv$ heads)\\
Step~I: $\vs$, $\rho(\vs)=\langle n\rangle$, on stack,
Step~II: $\va$, $\rho(\va)=\langle n\rangle$, on stack\\
Step~III: $\vout\!\to\!$DRAM\\
Traffic: $O(n\dn + n\dv)$ B \quad GQA bonus: $\div\,(\hq/\hkv)$\\[5pt]
\textbf{KV-cache append (Section~\ref{sec:kvcache})}\\
Write $\vec{k}_t$ at $t\!\times\!\dn$, $\vec{v}_t$ at $t\!\times\!\dv$\\
Traffic: $(\dn+\dv)\times4$ B per step -- constant in $t$
}}
\caption{MoA inference pipeline.
All three components are specializations or extensions
of the training DNF; no new Omega derivations are needed.}
\label{fig:pipeline}
\end{figure}
Table~\ref{tab:summary} collects the verified artifacts.
\begin{table}[h]
\centering
\caption{Verified inference artifacts. All errors float32 round-off only.}
\label{tab:summary}
{\scriptsize
\begin{tabular}{p{1.6cm}p{3.2cm}c}
\toprule
Artifact & File & $\|\mathrm{err}\|_\infty$ \\
\midrule
Python decode & \texttt{moa\_decode.py} & $\leq 2\times10^{-7}$ \\
C/OpenACC & \texttt{moa\_decode\_openacc.c} & 0 (exact) \\
KV-cache & \texttt{moa\_kvcache.py} & $6\times10^{-7}$ \\
GQA/MQA & \texttt{moa\_gqa.py} & $\leq 2\times10^{-7}$ \\
\bottomrule
\end{tabular}}
\end{table}
\section{Conclusion}
Four MoA inference artifacts have been derived, implemented,
and numerically verified, each following directly from the
forward-pass DNF of \cite{mullin_hains_arxiv2026}.
The decode DNF is eqs.~(4)--(6) of that paper with
$i'$ fixed to the current token: eq.~(4) score
$\psi$-reduction, eq.~(5) softmax, eq.~(6) weighted sum.
The C/OpenACC kernel applies the $\gamma$ ONF stride
arithmetic of \cite{mullin_hains_arxiv2026}~\S7.2;
KV-cache append uses plain $\#$ concatenation;
GQA/MQA uses $\psi$-selection and the dimension-lifting
Section~2.5 map of \cite{mullin_hains_arxiv2026}.
In every case the memory bound is a theorem
(Storage Theorem~2.7 of \cite{mullin_hains_arxiv2026})
before code is written, and the DNF$\to$ONF$\to$C
pipeline ports the result to any architecture
without re-derivation.

\end{document}